\documentclass[runningheads]{llncs}
\usepackage{graphicx}

\usepackage{tikz}
\usepackage{comment}
\usepackage{amsmath,amssymb} 
\usepackage{color}
\usepackage{subcaption}
\usepackage{multirow}
\usepackage{wrapfig}
\usepackage{threeparttable}

\usepackage[bookmarks=false]{hyperref}

\hypersetup{
  colorlinks,
  linkcolor=red,
  citecolor=black
}

\usepackage[accsupp]{axessibility}  

\begin{document}
\pagestyle{headings}
\mainmatter
\def\ECCVSubNumber{2516}  

\title{SparseNeuS: Fast Generalizable Neural Surface Reconstruction from Sparse Views} 

\titlerunning{Fast Generalizable Neural Surface Reconstruction from Sparse Views}
%
\author{Xiaoxiao Long\inst{1} \quad 
Cheng Lin\inst{2} \quad 
Peng Wang\inst{1} \\
Taku Komura\inst{1} \quad  Wenping Wang\inst{3}}
\authorrunning{X. Long et al.}
%
\institute{The University of Hong Kong\and Tencent Games\and Texas A\&M University
}


\newcommand{\Cheng}[1]{{\color{magenta} (Cheng: #1)}}
\newcommand{\wpeng}[1]{{\color{orange} (Peng: #1)}}

\maketitle


\definecolor{purple}{cmyk}{0.45,0.86,0,0}
\definecolor{bleudefrance}{rgb}{0.19, 0.55, 0.91}
\definecolor{darkorange}{rgb}{1, 0.55, 0}
\definecolor{limegreen}{rgb}{0.2, 0.8, 0.2}

\def\etal{et al.}			  
\def\eg{e.g.,~}               
\def\ie{i.e.,~}               
\def\etc{etc}                 
\def\cf{cf.~}                 
\def\viz{viz.~}               
\def\vs{vs.~}                 



\newlength\paramargin
\newlength\figmargin
\newlength\secmargin
\newlength\figcapmargin

\setlength{\secmargin}{0.0mm}
\setlength{\paramargin}{0.0mm}
\setlength{\figmargin}{0.0mm}
\setlength{\figcapmargin}{0.5mm}

\newcommand{\red}{\textcolor{red}}
\newcommand{\blue}{\textcolor{blue}}

\newcommand{\mpage}[2]
{
\begin{minipage}{#1\linewidth}\centering
#2
\end{minipage}
}

\newcommand{\mfigure}[2]
{
\begin{subfigure}[b]{#1\linewidth}\centering
\includegraphics[width=\linewidth]{#2}
\end{subfigure}
}

\newcommand{\Paragraph}[1]
{
\vspace{\paramargin}
\paragraph{#1}
}

\newcommand{\heading}[1]
{
\vspace{1mm}
\noindent \textbf{#1}
}   

\newcommand{\secref}[1]{Section~\ref{#1}}
\newcommand{\figref}[1]{Figure~\ref{#1}} 
\newcommand{\tblref}[1]{Table~\ref{#1}}
\newcommand{\eqnref}[1]{Equation~\ref{#1}}
\newcommand{\thmref}[1]{Theorem~\ref{#1}}
\newcommand{\prgref}[1]{Program~\ref{#1}}
\newcommand{\algref}[1]{Algorithm~\ref{#1}}
\newcommand{\clmref}[1]{Claim~\ref{#1}}
\newcommand{\lemref}[1]{Lemma~\ref{#1}}
\newcommand{\ptyref}[1]{Property~\ref{#1}}

\long\def\ignorethis#1{}
\newcommand {\todo}{{\textbf{\color{red}[TO-DO]\_}}}
\def\newtext#1{\textcolor{blue}{#1}}
\def\modtext#1{\textcolor{red}{#1}}

\newcommand{\LXX}[1]{{\color{cyan}          {[Xiao: #1]}}}
\newcommand{\ctc}[1]{{\color{darkorange}  {[CT: #1]}}}
\newcommand{\LC}[1]{{\color{blue}  {[LC: #1]}}}

\newcommand{\jbox}[2]{
  \fbox{%
  	\begin{minipage}{#1}%
  		\hfill\vspace{#2}%
  	\end{minipage}%
  }}

\newcommand{\jblock}[2]{%
	\begin{minipage}[t]{#1}\vspace{0cm}\centering%
	#2%
	\end{minipage}%
}

\begin{abstract}
We introduce {\em SparseNeuS}, a novel neural rendering based method for the task of surface reconstruction from multi-view images. 
This task becomes more difficult when only sparse images are provided as input, a scenario where existing neural reconstruction approaches usually produce incomplete or distorted results.
Moreover, their inability of generalizing to unseen new scenes impedes their application in practice. 
Contrarily, {\em SparseNeuS} can generalize to new scenes and work well with sparse images (as few as 2 or 3).
{\em SparseNeuS} adopts signed distance function (SDF) as the surface representation, and learns generalizable priors from image features by introducing \textit{geometry encoding} volumes for generic surface prediction.
Moreover, several strategies are introduced to effectively leverage sparse views for high-quality reconstruction, including 1) a multi-level geometry reasoning framework to recover the surfaces in a coarse-to-fine manner; 2) a multi-scale color blending scheme for more reliable color prediction; 3) a consistency-aware fine-tuning scheme to control the inconsistent regions caused by occlusion and noise. Extensive experiments demonstrate that our approach not only outperforms the state-of-the-art methods, but also exhibits good efficiency, generalizability, and flexibility\footnote[1]{ 
Visit our project page: \url{https://www.xxlong.site/SparseNeuS}}.

\keywords{reconstruction, volume rendering, sparse views}
\end{abstract}

\begin{figure}
    \setlength{\abovecaptionskip}{0pt}
    \setlength{\belowcaptionskip}{5pt}
    \centering
    \includegraphics[width=\textwidth]{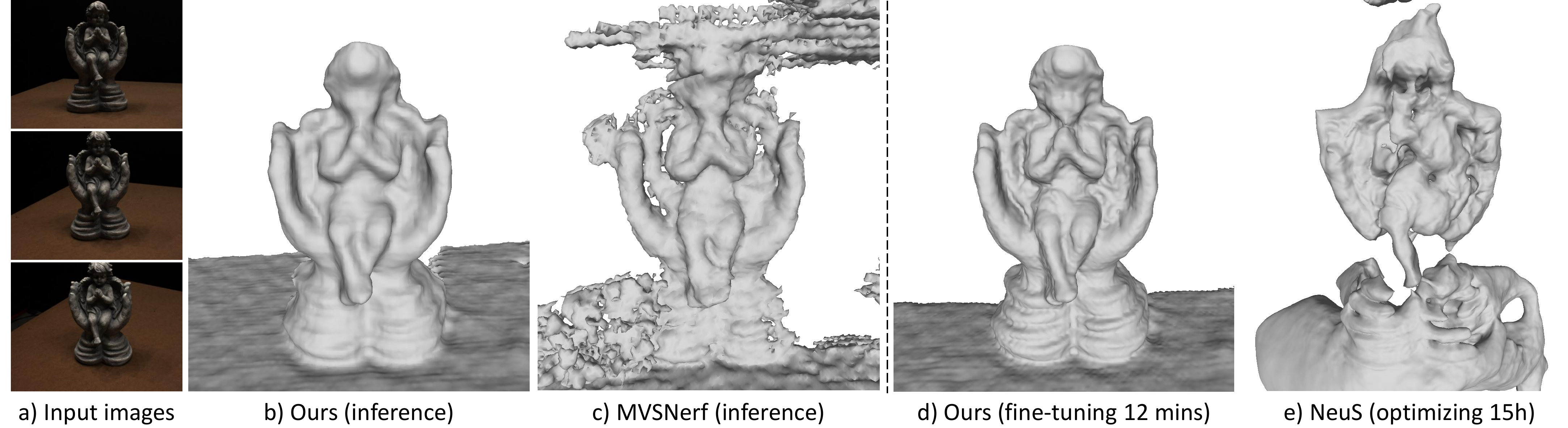}
    \caption{
    Our method can generalize across diverse scenes, and reconstruct neural surfaces from only three input images (a) via fast network inference (b). The reconstruction quality of the fast inference step is more accurate and faithful than the result of MVSNerf~\cite{chen2021mvsnerf} (c). Our inference result can be further improved by a per-scene fine-tuning process. Compared to NeuS~\cite{wang2021neus} (e), our per-scene optimization result (d) not only achieves noticeably better reconstruction quality, but also takes much less time to converge (12 minutes v.s. 15 hours).
    }
    \label{fig:teaser}
\end{figure}

\section{Introduction}

Reconstructing 3D geometry from multi-view images is a fundamental problem in computer vision and has been extensively researched for decades. 
Conventional methods for multi-view stereo~\cite{campbell2008using,galliani2015massively,schonberger2016pixelwise,kutulakos2000theory,seitz1999photorealistic,furukawa2009accurate,lhuillier2005quasi} reconstruct 3D geometry from input images by finding corresponding matches across the input images. However, when only a sparse set of images are available as input, image noises, weak textures and reflections make it difficult for these methods to build dense and complete matches.

With the recent advances in neural implicit representations, neural surface reconstruction methods~\cite{wang2021neus,yariv2020multiview,yariv2021volume,oechsle2021unisurf} leverage neural rendering to jointly optimize the implicit geometry and the radiance field by minimizing the difference of rendered views and ground truth views. Although the methods can produce plausible geometry and photorealistic novel views, they suffer from two major limitations. First, existing methods heavily depend on a large number of input views, i.e. dense views, that are often not available in practice. Second, they require time-consuming per-scene optimization for reconstruction, thus incapable of generalizing to new scenes. The limitations need to be resolved for making such reconstruction methods relevant and useful for practical application.

We propose {\em SparseNeuS}, a novel multi-view surface reconstruction method with two distinct advantages: 1) it generalizes well to new scenes; 2) it needs only a sparse set of images (as few as 2 or 3 images) for successful reconstruction. {\em SparseNeuS} achieves these goals by learning generalizable priors from image features and hierarchically leverages the information encoded in the sparse input. 

To learn generalizable priors, following MVSNerf~\cite{chen2021mvsnerf}, we construct a \textit{geometry encoding} volume which aggregates the 2D image features from multi-view input, and use these informative latent features to infer 3D geometry. Consequently, our surface prediction network takes a hybrid representation as input, i.e., $xyz$ coordinates and the corresponding features from the geometry encoding volume, to predict the network-encoded signed distance function (SDF) for the reconstructed surface. 

The most crucial part of our pipeline is in how to effectively incorporate the limited information from sparse input images to obtain high-quality surfaces through neural rendering. To this end, we introduce several strategies to tackle this challenge. The first is a {\em multi-level geometry reasoning scheme} to progressively construct the surface from coarse to fine. We use a cascaded volume encoding structure, i.e., a coarse volume that encodes relatively global features to obtain the high-level geometry, and a fine volume guided by the coarse level to refine the geometry. A per-scene fine-tuning process is further incorporated into this scheme, which is conditioned on the inferred geometry to construct subtle details to generate even finer-grained surfaces.
This multi-level scheme divides the task of high-quality reconstruction into several steps. Each step is based upon the geometry from the preceding step and focuses on constructing a finer level of details. Besides, due to the hierarchical nature of the scheme, the reconstruction efficiency is significantly boosted, because numerous samples far from the coarse surface can be discarded, so as not to burden the computation in the fine-level geometry reasoning.

The second important strategy that we propose is a {\em multi-scale color bending scheme} for novel view synthesis. Given the limited information in the sparse images, the network would struggle to directly regress accurate colors for rendering novel views. Thus, we mitigate this issue by predicting the linear blending weights of the input image pixels to derive colors. Specifically, we adopt both pixel-based and patch-based blending to jointly evaluate local and contextual radiance consistency. This multi-scale blending scheme yields more reliable color predictions when the input is sparse.

Another challenge in multi-view 3D reconstruction is that 3D surface points often do not have consistent projections across different views, due to occlusion or image noises. With only a small number of input views, the dependence of geometry reasoning on each image further increases, which aggravates the problem and results in distorted geometry. To tackle this challenge, we propose a {\em consistency-aware fine-tuning scheme} in the fine-tuning stage. This scheme automatically detects regions that lack consistent projections, and excludes these regions in the optimization. This strategy proves effective in making the fine-tuned surface less susceptible to occlusion and noises, thus more accurate and cleaner, contributing to a high-quality reconstruction.

We evaluated our method on the DTU~\cite{jensen2014large} and BlendedMVS~\cite{yao2020blendedmvs} datasets, and show that our method outperforms the state-of-the-art unsupervised neural implicit surface reconstruction methods both quantitatively and qualitatively. 

In summary, our main contributions are:
\begin{itemize}
    \item We propose a new surface reconstruction method based on neural rendering. Our method learns generalizable priors across scenes and thus can generalize to new scenes for 3D reconstruction with high-quality geometry.
    
    \item  Our method is capable of high-quality reconstruction from a sparse set of images, as few as 2 or 3 images. This is achieved by effectively inferring 3D surfaces from sparse input images using three novel strategies: a) multi-level geometry reasoning; b) multi-scale color blending; and c) consistency-aware fine-tuning.

    \item Our method outperforms the state-of-the-arts in both reconstruction quality and computational efficiency. 
\end{itemize}



\section{Related Work}

\subsection{Multi-view stereo (MVS)}
Classical MVS methods utilize various 3D representations for reconstruction such as: 
voxel grids based~\cite{ji2017surfacenet,ji2020surfacenet+,kar2017learning,kutulakos2000theory,seitz1999photorealistic,sun2021neuralrecon},
3D point clouds based~\cite{furukawa2009accurate,lhuillier2005quasi},
and depth maps based~\cite{campbell2008using,galliani2015massively,schonberger2016pixelwise,tola2012efficient,yao2018mvsnet,yao2019recurrent,gu2020cascade,long2020occlusion,long2021multi,long2021adaptive}.
Compared with voxel grids and 3D point clouds, depth maps are much more flexible and appropriate for parallel computation, so depth map based methods are most common, like the well-known method COLMAP~\cite{schonberger2016pixelwise}.
Depth map based methods first estimate the depth map of each image, and then utilize filtering operations to fuse the depth maps together into a global point cloud, which can be further processed using a meshing algorithm like Screened Poisson surface reconstruction~\cite{kazhdan2013screened}.
These methods achieve promising results with densely captured images.
However, with a limited number of images, these methods become more sensitive to image noises, weak textures and reflections, making it difficult for these methods to produce complete reconstructions.

\subsection{Neural surface reconstruction}
Recently, neural implicit representations of 3D geometry are successfully applied in shape modeling~\cite{atzmon2020sal,chen2019learning,gropp2020implicit,mescheder2019occupancy,park2019deepsdf,michalkiewicz2019implicit}, novel view synthesis~\cite{sitzmann2019scene,lombardi2019neural,mildenhall2020nerf,liu2020neural,saito2019pifu,sitzmann2019deepvoxels,sitzmann2019deepvoxels,trevithick2021grf} and mutli-view 3D reconstruction~\cite{jiang2020sdfdiff,yariv2020multiview,niemeyer2020differentiable,kellnhofer2021neural,liu2020dist,wang2021neus,yariv2021volume,oechsle2021unisurf,zhang2021learning,darmon2021improving}.
For the task of multi-view reconstruction, the 3D geometry is represented by a neural network which outputs either occupancy field or Signed Distance Function (SDF).
Some methods utilize surface rendering~\cite{niemeyer2020differentiable} for multi-view reconstruction, but they always need extra object masks~\cite{yariv2020multiview,niemeyer2020differentiable} or depth priors~\cite{zhang2021learning}, which is inefficient for practical applications.
To avoid extra masks or depth priors, some methods~\cite{wang2021neus,yariv2021volume,oechsle2021unisurf,darmon2021improving} leverage volume rendering for reconstruction.
However, they also heavily depend on a large number of images to perform a time-consuming per-scene optimization, thus incapable of generalizing to new scenes.


In terms of generalization, there are some successful attempts~\cite{yu2021pixelnerf,wang2021ibrnet,chen2021mvsnerf,liu2021neural,chibane2021stereo} for neural rendering. These methods take sparse views as input and make use of the radiance information of the images to generate novel views, and can generalize to unseen scenes. 
Although they can generate plausible synthesized images, the extracted geometries from these methods always suffer from noises, incompleteness and distortion.

\begin{figure}[t]
    \setlength{\abovecaptionskip}{0pt}
    \setlength{\belowcaptionskip}{5pt}
    \centering
    \includegraphics[width=\textwidth]{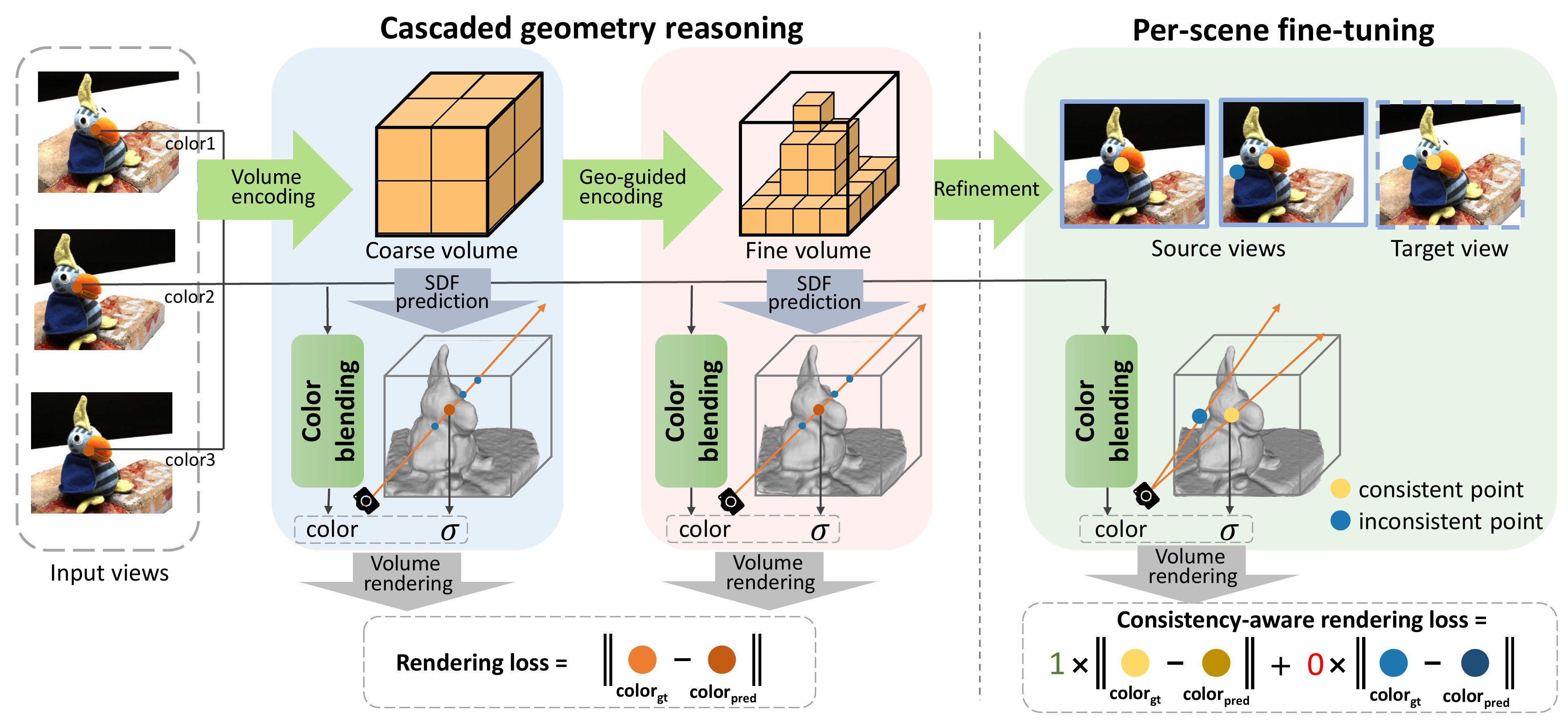}
    \caption{
    The overview of {\em SparseNeuS}. 
    The cascaded geometry reasoning scheme first constructs a coarse volume that encodes relatively global features to obtain the fundamental geometry, and then constructs a fine volume guided by the coarse level to refine the geometry. Finally, a consistency-aware fine-tuning strategy is used to add subtle geometry details, thus yielding high-quality reconstructions with fine-grained surfaces. Specially, a multi-scale color blending module is leveraged for more reliable color prediction.
    }
    \label{fig:pipeline}
\end{figure}

\section{Method}

Given a few (i.e., three) views with known camera parameters, we present a novel method that hierarchically recovers surfaces and generalizes across scenes. 
As illustrated in Figure~\ref{fig:pipeline}, our pipeline can be divided into three parts:
(1) \textbf{Geometry reasoning}. {\em SparseNeuS} first constructs cascaded \textit{geometry encoding} volumes that encode local geometry surface information, and recover surfaces from the volumes in a coarse-to-fine manner (see Section~\ref{geometry_reasoning}).
(2) \textbf{Appearance prediction}.
{\em SparseNeuS} leverages a multi-scale color blending module to predict colors by aggregating information from input images, and then combines the estimated geometry with predicted colors to render synthesized views using volume rendering (see Section~\ref{appearance_prediction}).
(3) \textbf{Per-scene fine-tuning}.
Finally, a consistency-aware fine-tuning scheme is proposed to further improve the obtained geometry with fine-grained details (see Section~\ref{fine-tuning}).

\subsection{Geometry reasoning}
\label{geometry_reasoning}
{\em SparseNeuS} constructs cascaded \textit{geometry encoding} volumes of two different resolutions for geometry reasoning, which aggregates image features to encode the information of local geometry. Specially, the coarse geometry is first extracted from a \textit{geometry encoding} volume of low resolution, and then it is used to guide the geometry reasoning of the fine level.

\noindent
\textbf{Geometry encoding volume.}
For the scene captured by $N$ input images ${\{I_i\}}_{i=0}^{N-1}$, we first estimate a bounding box which can cover the region of interests.
The bounding box is defined in the camera coordinate system of the centered input image, and then grided into regular voxels.
To construct a \textit{geometry encoding} volume $M$, 2D feature maps ${\{ F_i\}}_{i=0}^{N-1}$ are extracted from the input images ${\{I_i\}}_{i=0}^{N-1}$ by a 2D feature extraction network.
Next, with the camera parameters of one image $I_i$, we project each vertex $v$ of the bounding box to each feature map $F_i$ and obtain its features $F_i(\pi_{i}(v))$ by interpolation, where $\pi_{i}(v)$ denotes the projected pixel location of $v$ on the feature map $F_i$. For simplicity, we abbreviate $F_i(\pi_{i}(v))$ as $F_{i}(v)$.

The \textit{geometry encoding} volume $M$ is constructed using all the projected features $ {\{F_{i}(v)\}}_{i=0}^{N-1}$ of each vertex. Following prior methods~\cite{yao2018mvsnet,chen2021mvsnerf}, we first calculate the variance of all the projected features of a vertex to build a cost volume $B$, and then apply a sparse 3D CNN $\Psi$ to aggregate the cost volume $B$ to obtain the \textit{geometry encoding} volume $M$:
\begin{equation}
M=\psi (B), \quad  B(v)= \operatorname{Var}\left(
{\{F_{i}(v)\}}_{i=0}^{N-1}
\right),
\end{equation}
where $\operatorname{Var}$ is the variance operation, which computes the variance of all the projected features ${\{F_{i}(v)\}}_{i=0}^{N-1}$ of each vertex $v$.

\noindent
\textbf{Surface extraction.}
Given an arbitrary 3D location $q$, an MLP network $f_{\theta}$ takes the combination of the 3D coordinate and its corresponding interpolated features of \textit{geometry encoding} volume $M(q)$ as input, to predict the Signed Distance Function (SDF) $s(q)$ for surface representation.
Specially, positional encoding $\operatorname{PE}$ is applied on its 3D coordinates, and the surface extraction operation is expressed as:
$s(q) = f_{\theta}(\operatorname{PE}(q), M(q)).$

\noindent
\textbf{Cascaded volumes scheme.}
For balancing the computational efficiency and reconstruction accuracy, {\em SparseNeuS} constructs cascaded \textit{geometry encoding} volumes of two resolutions to perform geometry reasoning in a coarse-to-fine manner.
A coarse \textit{geometry encoding} volume is first constructed to infer the fundamental geometry, which presents the global structure of the scene but is relatively less accurate due to limited volume resolution.
Guided by the obtained coarse geometry, a fine level \textit{geometry encoding} volume is constructed to further refine the surface details.
Numerous vertices far from the coarse surfaces can be discarded in the fine-level volume, which significantly reduces the computational memory burden and improves efficiency. 

\subsection{Appearance prediction}
\label{appearance_prediction}
Given an arbitrary 3D location $q$ on a ray with direction $d$, we predict its color by aggregating appearance information from the input images.
Given limited information in the sparse input images, it is difficult for a network to directly regress color values for rendering novel views. Unlike prior works~\cite{yu2021pixelnerf,chen2021mvsnerf}, {\em SparseNeuS} predicts blending weights of the input images to generate new colors.
A location $q$ is first projected to the input images to obtain the corresponding colors ${\{I_i(q)\}}_{i=0}^{N-1}$. Then the colors from different views are blended together as the predicted color of $q$ using the estimated blending weights.

\noindent
\textbf{Blending weights.} The key of generating the blending weights ${\{w_{i}^{q} \}}_{i=0}^{N-1}$ is to consider the photography consistency of the input images.
We project $q$ onto the feature maps ${\{ F_i\}}_{i=0}^{N-1}$ to extract the corresponding features ${\{ F_i(q)\}}_{i=0}^{N-1}$ using bilinear interpolation.
Moreover, we calculate the mean and variance of the features ${\{ F_i(q)\}}_{i=0}^{N-1}$ from different views to capture the global photographic consistency information.
Each feature $F_{i}(q)$ is concatenated with the mean and variance together, and then fed into a tiny MLP network to generate a new feature $F_{i}^{\prime}(q)$.
Next, we feed the new feature $F_i^{\prime}(q)$, 
the viewing direction of the query ray relative to the viewing direction of the $i_{th}$ input image $ \Delta d_i = d - d_i$, and the trilinearly interpolated volume encoding feature $M(q)$ into an MLP network $f_{c}$ to generate blending weight:
$w_{i}^{q}=f_{c}(F_{i}^{\prime}(q),M(q), \Delta d_i).$
Finally, blending weights ${\{w_{i}^{q} \}}_{i=0}^{N-1}$ are normalized using a Softmax operator.

\noindent
\textbf{Pixel-based color blending.}
With the obtained blending weights, the color $c_q$ of a 3D location $q$ is predicted as the weighted sum of its projected colors ${\{I_i(q)\}}_{i=0}^{N-1}$ on the input images.
To render the color of the query ray, we first predict the color and SDF values of 3D points sampled on the ray.
The color and SDF values of the sampled points are aggregated to obtain the final colors of the ray using SDF based volume rendering~\cite{wang2021neus}.
Since the color of a query ray corresponds to a pixel of the synthesized image, we name this operation pixel-based blending.
Although supervision on the colors rendered by pixel-based blending already induces effective geometry reasoning, the information of a pixel is local and lacks contextual information, thus usually leading to inconsistent surface patches when input is sparse.

\noindent
\textbf{Patch-based color blending.} 
Inspired by classical patch matching, we consider
enforcing the synthesized colors and ground truth colors to be contextually consistent; that is, not only in pixel level but also in patch level.
To render the colors of a patch with size $k \times k$, a naive implementation is to query the colors of $k^2$ rays using volume rendering, which causes a huge amount of computation. We, therefore, leverage local surface plane assumption and homography transformation to achieve a more efficient implementation.

The key idea is to estimate a local plane of a sampled point to efficiently derive the local patch. Given a sampled point $q$, we leverage the property of the SDF network $s(q)$ to estimate the normal direction $n_q$ by computing the spatial gradient, i.e., $n_q=\nabla s(q)$. Then, we sample a set of points on the local plane $(q, n_q)$, project the sampled points to each view, and obtain the colors by interpolation on each input image.
All the points on the local plane share the same blending weights with $q$, and thus only one query of the blending weights is needed.
Using local plane assumption, we consider the neighboring geometric information of a query 3D position, which encodes contextual information of local patches and enforces better geometric consistency. By adopting patch-based volume rendering, synthesized regions contain more global information than single pixels, thus producing more informative and consistent shape context, especially in the regions with weak texture and changing intensity.

\noindent
\textbf{Volume rendering.} To rendering the pixel-based color $C\left( r \right)$ or patch-based color $P\left( r \right)$ of a ray $r$ passing through the scene, we query the pixel-based colors $c_i$, patch-based colors $p_i$ and sdf values $s_i$ of $M$ samples on the ray, and then utilize~\cite{wang2021neus} to convert sdf values $s_i$ into densities $\sigma_{i}$. Finally, the densities are used to accumulate pixel-based and patch-based colors along the ray:

\begin{equation}
\label{eq_volume_rendering}
U(r)=\sum_{i=1}^{M} T_{i}\left(1-\exp \left(-\sigma_{i}\right)\right) u_{i}, \quad \text{where} \quad T_{i}=\exp \left(-\sum_{j=1}^{i-1} \sigma_{j}\right).                                             
\end{equation}
Here $U(r)$ denotes $C\left( r \right)$ or $P\left( r \right)$, while $u_i$ denotes the pixel-based color  $c_i$ or patch-based color $p_i$ of the $i_{th}$ sample on the ray.

\subsection{Per-scene fine-tuning}
\label{fine-tuning}
With the generalizable priors and effective geometry reasoning framework, given sparse images from a new scene, {\em SparseNeuS} can already recover geometry surfaces via fast network inference. 
However, due to the limited information in the sparse input views and the high diversity and complexity of different scenes, the geometry obtained by the generic model may contain inaccurate outliers and lack subtle details. 
Therefore, we propose a novel fine-tuning scheme, which is conditioned on the inferred geometry, to reconstruct subtle details and generate finer-grained surfaces. 
Thanks to the initialization given by the network inference, the per-scene optimization can fast converge to a high-quality surface.

\noindent
\textbf{Fine-tuning networks.}
In the fine-tuning, we directly optimize the obtained fine-level \textit{geometry encoding} volume and the signed distance function (SDF) network $f_{\theta}$, while the 2D feature extraction network and 3D sparse CNN networks are discarded.
Moreover, the CNN based blending network used in the generic setting is replaced by a tiny MLP network.
Although the CNN based network can be also used in per-scene fine-tuning, by experiments, we found that a new tiny MLP can speed up the fine-tuning without loss of performance since the MLP is much smaller than the CNN based network.
The MLP network still outputs blending weights ${\{w_{i}^q \}}_{i=0}^{N-1}$ of a query 3D position $q$, but it takes the input as the combination of 3D coordinate $q$, the surface normal $n_q$, the ray direction $d$, the predicted SDF $s(q)$, and the interpolated feature of the \textit{geometry encoding} volume $M(q)$. Specially, positional encoding $\operatorname{PE}$ is applied on the 3D position $q$ and the ray direction $d$. The MLP network $f_{c}^{\prime}$ is defined as :
$
    {\{w_{i}^q \}}_{i=0}^{N-1} = f_{c}^{\prime}\left(\operatorname{PE}(q), 
    \operatorname{PE}(d),n_q,s(q),M(q)
    \right),
$
where ${\{w_{i}^q \}}_{i=0}^{N-1}$ are the predicted blending weights, and $N$ is the number of input images.

\noindent
\textbf{Consistency-aware color loss.}
We observe that in multi-view stereo, 3D surface points often do not have consistent projections across different views, since the projections may be occluded or contaminated by image noises. 
As a result, the errors of these regions suffer from sub-optima,
and the predicted surfaces of the regions are always inaccurate and distorted.
To tackle this problem, we propose a consistency-aware color loss to automatically detect the regions lacking consistent projections and exclude these regions in the optimization:

\begin{equation}
\label{ca_color_loss}
\resizebox{0.9\hsize}{!}{%
$ 
\begin{split}
    \mathcal{L}_{color} & =
    \sum_{r \in \mathbb{R}}  
    O \left(r \right) \cdot \mathcal{D}_{pix}\left(C \left(r \right), \tilde{C}\left(r \right)\right)
      + \sum_{r \in \mathbb{R}}  
      O \left(r \right) \cdot \mathcal{D}_{pat}\left(P\left(r \right),\tilde{P}\left(r \right)\right) \\
     & + \lambda_{0} \sum_{r \in \mathbb{R}} log\left(O\left(r \right)\right) 
      + \lambda_{1} \sum_{r \in \mathbb{R}} log\left(1- O\left(r \right)\right),
\end{split}
$ 
}
\end{equation}
where $ r$ is a query ray, $\mathbb{R}$ is the set of all query rays, $O \left(r \right)$ is the sum of accumulated weights along the ray $ r$ obtained by volume rendering. From Eq.~\ref{eq_volume_rendering}, we can easily derive
$O \left(r \right) = \sum_{i=1}^{M} T_{i}\left(1-\exp \left(-\sigma_{i}\right)\right)$.
$C \left(r \right)$ and $\tilde{C}\left(r \right)$ are the rendered and ground truth pixel-based colors of the query ray respectively, $P \left(r \right)$ and $\tilde{P}\left(r \right)$ are the rendered and ground truth patch-based colors of the query ray respectively, and
$\mathcal{D}_{pix}$ and $\mathcal{D}_{pat}$ are the loss metrics of the rendered pixel color and rendered patch colors respectively. Empirically, we choose $\mathcal{D}_{pix}$ as L1 loss and $\mathcal{D}_{pat}$ as Normalized Cross Correlation (NCC) loss.

The rationale behind this formulation is, the points with inconsistent projections always have relatively large color errors that cannot be minimized in the optimization. Therefore, if the color errors are difficult to be minimized in optimization, we force the sum of the accumulated weights $O \left(r \right)$ to be zero, such that the inconsistent regions will be excluded in the optimization. To control the level of consistency, we introduce two logistic regularization terms: decreasing the ratio $\lambda_{0}/\lambda_{1}$ will lead to more regions being kept; otherwise, more regions are excluded and the surfaces are cleaner.

\subsection{Training loss}
By enforcing the consistency of the synthesized colors and ground truth colors, the training of {\em SparseNeuS} does not rely on 3D ground-truth shapes. The overall loss function is defined as a weighted sum of the three loss terms: 
\begin{equation}
\label{total_loss}
\mathcal{L}= \mathcal{L}_{\text {color }}+\alpha \mathcal{L}_{\text {eik}} + \beta \mathcal{L}_{\text {sparse}}.
\end{equation}


We note that, in the early stage of generic training, the estimated geometry is relatively inaccurate, and 3D surface points may have large errors, where the errors do not provide clear clues on whether the regions are radiance consistent or not. We utilize consistency-aware color loss in the per-scene fine-tuning, and remove the last two consistence-aware logistic terms of Eq.~\ref{ca_color_loss} in the training of the generic model.

An Eikonal term~\cite{gropp2020implicit} is applied on the sampled points to regularize the SDF values derived from the surface prediction network $f_{\theta}$:
\begin{equation}
\mathcal{L}_{eik}=\frac{1}{\left\|\mathbb{Q} \right\|} \sum_{q \in \mathbb{Q}}\left({\left\|\nabla f_{\theta}\left(q\right)\right\|}_{2}-1\right)^{2},
\end{equation}
where $q$ is a sampled 3D point, $\mathbb{Q}$ is the set of all sampled points, $\nabla f_{\theta}\left(q\right)$ is the gradient of network $f_{\theta}$ relatively to sampled point $q$, and ${\left\| \cdot \right\|}_{2}$ is $l_2$ norm. The Eikonal term enforces the network $f_{\theta}$ to have unit $l_2$ norm gradient, which encourages $f_{\theta}$ to generate smooth surfaces.

Besides, due to the property of accumulated transmittance in volume rendering, the invisible query samples behind the visible surfaces lack supervision, which causes uncontrollable free surfaces behind the visible surfaces. To enable our framework to generate compact geometry surfaces, we adopt a sparseness regularization term to penalize the uncontrollable free surfaces:
\begin{equation}
\label{sparse_lossterm}
    \mathcal{L}_{sparse}=\frac{1}{\left\|\mathbb{Q} \right\| } \sum_{q \in \mathbb{Q}} \exp \left(-\tau \cdot \left|s(q) \right|\right),
\end{equation}
where $\left|s(q) \right|$ is the absolute SDF value of sampled point $q$, $\tau$ is a hyperparamter to rescale the SDF value. This term will encourage the SDF values of the points behind the visible surfaces to be far from 0. When extracting 0-level set SDF to generate mesh, this term can avoid uncontrollable free surfaces.




\section{Datasets and Implementation}
\begin{table*}[t]
\begin{center}
\caption{Evaluation on DTU~\cite{jensen2014large} dataset.}
\resizebox{\linewidth}{!}{%
\begin{threeparttable}[b]
\begin{tabular}{l|ccccc ccccc ccccc | c}
\hline
Scan &24 & 37 & 40 & 55 & 63 & 65 & 69 & 83 & 97 & 105 & 106 & 110 & 114 & 118 & 122 & Mean \\
\hline
PixelNerf~\cite{yu2021pixelnerf} & 5.13 & 8.07 & 5.85 & 4.40 & 7.11 & 4.64 & 5.68 & 6.76 & 9.05 & 6.11 & 3.95 & 5.92 & 6.26 & 6.89 & 6.93 & 6.28 \\
IBRNet~\cite{wang2021ibrnet} & 2.29 & 3.70 & 2.66 & 1.83 & 3.02 & 2.83 & 1.77 & 2.28 & 2.73 & 1.96 & 1.87 & 2.13 & 1.58 & 2.05 & 2.09 & 2.32 \\
MVSNerf~\cite{chen2021mvsnerf} & 1.96 & 3.27 & 2.54 & 1.93 & 2.57 & 2.71 & 1.82 & 1.72 & 2.29 & 1.75 & 1.72 & 1.47 & 1.29 & 2.09 & 2.26 & 2.09 \\
Ours & \textbf{1.68} & \textbf{3.06} & \textbf{2.25} & \textbf{1.10} & \textbf{2.37} & \textbf{2.18} & \textbf{1.28} & \textbf{1.47} & \textbf{1.80} & \textbf{1.23} & \textbf{1.19} & \textbf{1.17} & \textbf{0.75} & \textbf{1.56} & \textbf{1.55} & \textbf{1.64} \\
\hline
IDR\tnote{$\dagger$}~\cite{yariv2020multiview} & 4.01 & 6.40 & 3.52 & 1.91 & 3.96 & 2.36 & 4.85 & 1.62 & 6.37 & 5.97 & 1.23 & 4.73 & 0.91 & 1.72 & 1.26 & 3.39 \\
VolSdf~\cite{yariv2021volume} & 4.03 & 4.21 & 6.12 & 0.91 & 8.24 & 1.73 & 2.74 & 1.82 & 5.14 & 3.09 & 2.08 & 4.81 & 0.60 & 3.51 & 2.18 & 3.41 \\
UniSurf~\cite{oechsle2021unisurf} & 5.08 & 7.18 & 3.96 & 5.30 & 4.61 & 2.24 & 3.94 & 3.14 & 5.63 & 3.40 & 5.09 & 6.38 & 2.98 & 4.05 & 2.81 & 4.39 \\
Neus~\cite{wang2021neus} & 4.57 & 4.49 & 3.97 & 4.32 & 4.63 & 1.95 & 4.68 & 3.83 & 4.15 & 2.50 & 1.52 & 6.47 & 1.26 & 5.57 & 6.11 & 4.00 \\
IBRNet-ft~\cite{wang2021ibrnet} & 1.67 & 2.97 & 2.26 & 1.56 & 2.52 & 2.30 & 1.50 & 2.05 & 2.02 & 1.73 & 1.66 & 1.63 & 1.17 & 1.84 & 1.61 & 1.90 \\
Colmap~\cite{schonberger2016structure} & \textbf{0.90} & 2.89 & 1.63 & 1.08 & 2.18 & 1.94 & 1.61 & 1.30 & 2.34 & 1.28 & 1.10 & 1.42 & 0.76 & 1.17 & \textbf{1.14} & 1.52 \\
Ours-ft & 1.29 & \textbf{2.27} & \textbf{1.57} & \textbf{0.88} & \textbf{1.61} & \textbf{1.86} & \textbf{1.06} & \textbf{1.27} & \textbf{1.42} & \textbf{1.07} & \textbf{0.99} & \textbf{0.87} & \textbf{0.54} & \textbf{1.15} & 1.18 & \textbf{1.27} \\
\hline
\end{tabular}%

\begin{tablenotes}
     \item[$\dagger$] Optimization using extra object masks.
   \end{tablenotes}
\end{threeparttable}
}
\label{tab: dtu_eval}
\end{center}
\end{table*}
\noindent
\textbf{Datasets.} We train our framework on the DTU~\cite{jensen2014large} dataset to learn a generalizable network. 
We use 15 scenes for testing, same as those used in IDR~\cite{yariv2020multiview}, and the remaining non-overlapping 75 scenes for training. All the evaluation results on the testing scenes are generated using three views with a resolution of $600 \times 800$, and each scene contains two sets of three images. 
The foreground masks provided by IDR~\cite{yariv2020multiview} are used for evaluating the testing scenes. 
For memory efficiency, we use the center cropped images with resolution of $512 \times 640$ for training.
We observe that the images of DTU dataset contain large black backgrounds and the regions have considerable image noises, so we utilize a simple threshold based denoising strategy to alleviate the noises of such regions in the training images.
Optionally, the black backgrounds with zero RGB values can be used as a simple dataset prior to encourage the geometry predictions of such regions to be empty.
We further tested on 7 challenging scenes from the BlendedMVS~\cite{yao2020blendedmvs} dataset. For each scene, we select one set of three images with a resolution of $768 \times 576$ as input.
Note that, in the per-scene fine-tuning stage, we still use the three images for optimization without any new images.

\noindent
\textbf{Implementation details.}
Feature Pyramid Network~\cite{lin2017feature} is used as the image feature extraction network to extract multi-scale features from input images.
We implement the sparse 3D CNN networks using a U-Net like architecture, and use torchsparse~\cite{tang2020searching} as the implementation of 3D sparse convolution.
The resolutions of the coarse level and fine level \textit{geometry encoding} volumes are $96 \times 96 \times 96$ and $192 \times 192 \times 192$ respectively. 
The patch size used in patch-based blending is $5 \times 5$.
We adopt a two-stage training strategy to train our generic model: in the first stage, the networks of coarse level are first trained for 150k iterations;
in the second stage, the networks of fine level are trained for another 150k iterations while the networks of coarse level are fixed.
We train our model on two RTX 2080Ti GPUs with a batch size of 512 rays.

\section{Experiments}

\begin{figure}[!t]
    \centering
  \includegraphics[width=0.9\textwidth]{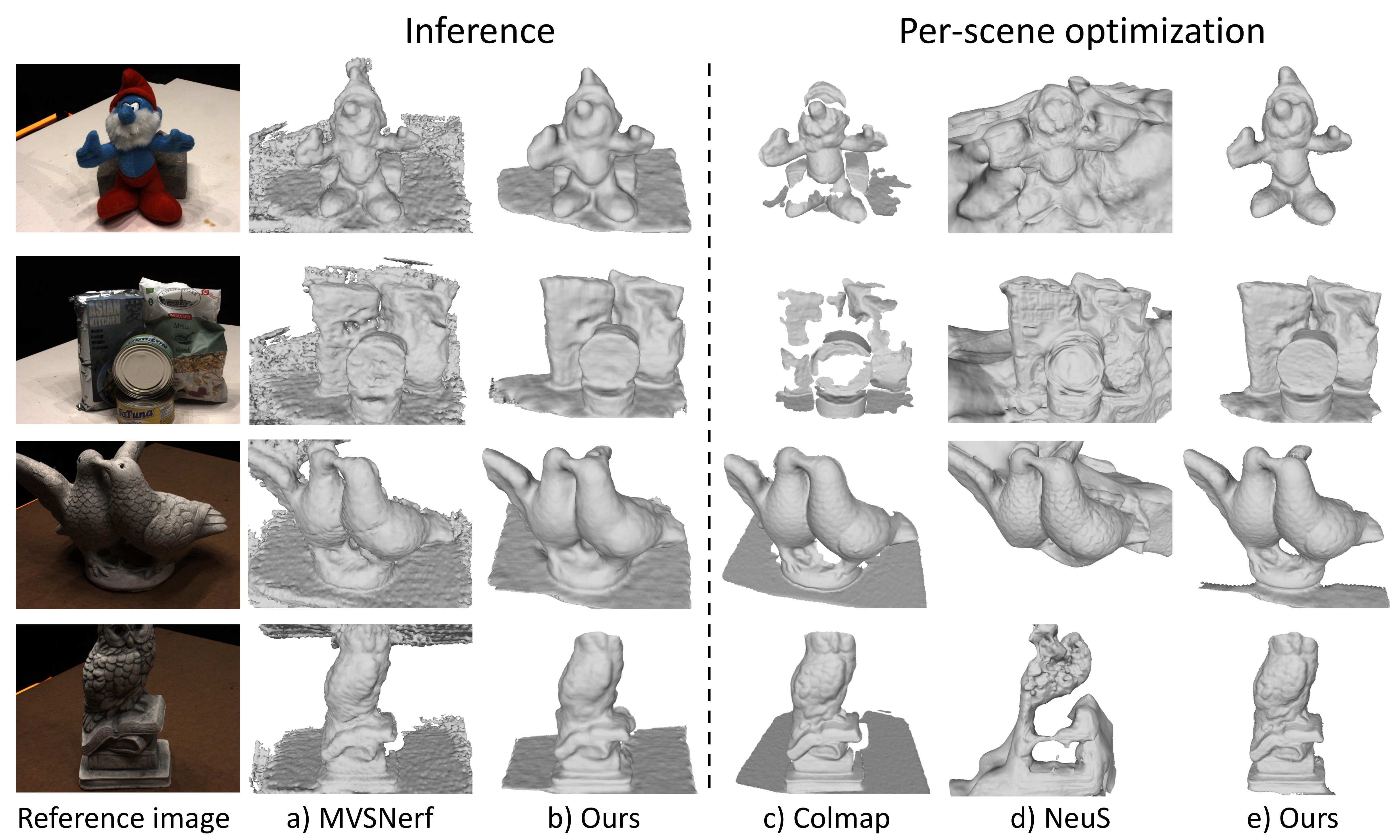}
    \caption{Visual comparisons on DTU~\cite{jensen2014large} dataset.}
    \label{fig:dtu_compare}
\end{figure}

We compare our method with the state-of-the-art approaches from three classes: 1) generic neural rendering methods, including PixelNerf~\cite{yu2021pixelnerf}, IBRNet~\cite{wang2021ibrnet} and MVSNerf~\cite{chen2021mvsnerf}, where we use a density threshold to extract meshes from the learned implicit field; 2) per-scene optimization based neural surface reconstruction methods, including IDR~\cite{yariv2020multiview}, NeuS~\cite{wang2021neus}, VolSDF~\cite{yariv2021volume}, and UniSurf~\cite{oechsle2021unisurf};
3) a widely used classic MVS method COLMAP~\cite{schonberger2016structure}, where we reconstruct a mesh from the output point cloud of COLMAP with Screened
Poisson Surface Reconstruction~\cite{kazhdan2013screened}. All the methods take three images as input.

\begin{figure}[!t]
    \centering
  \includegraphics[width=0.9\textwidth]{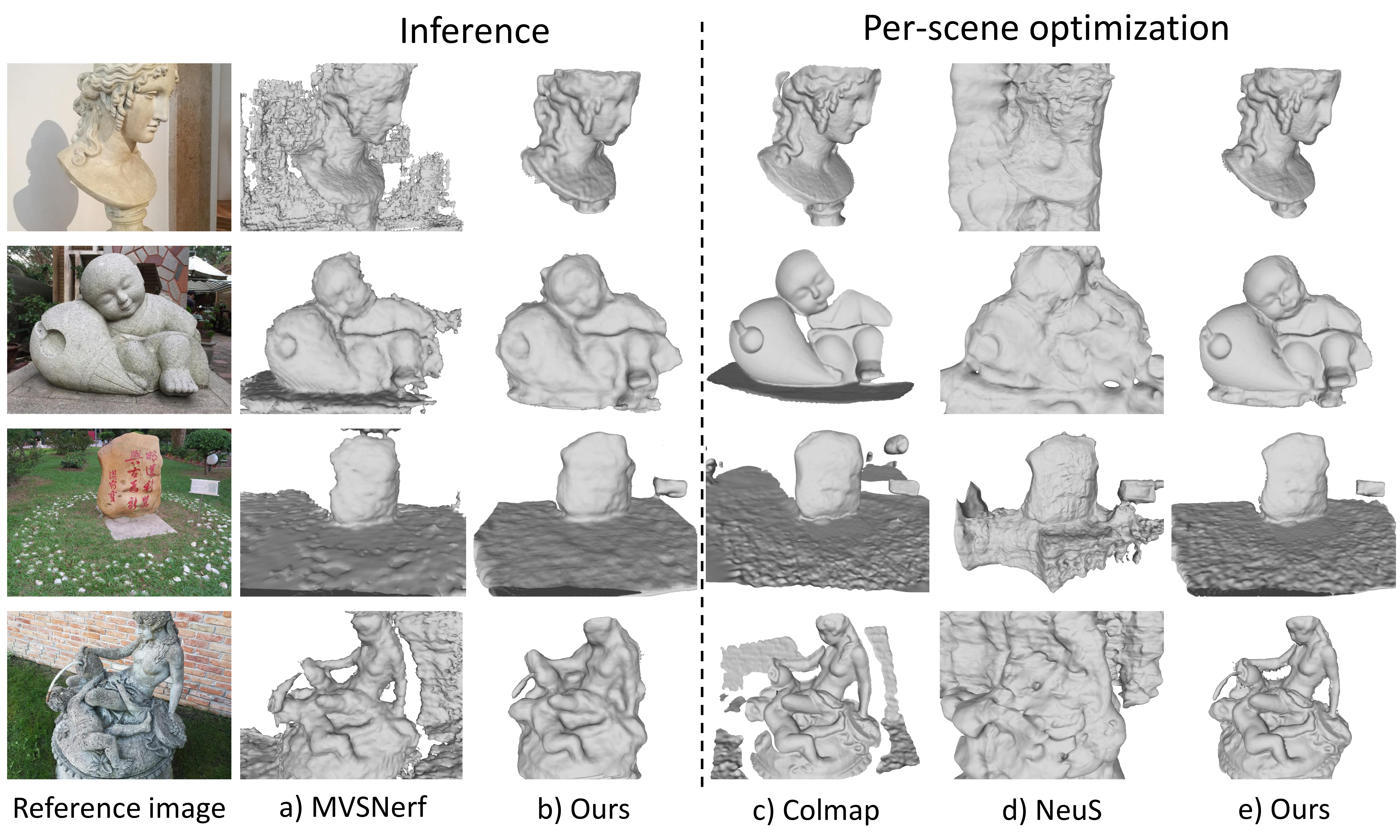}
    \caption{Visual comparisons on BlendedMVS~\cite{yao2020blendedmvs} dataset.}
    \label{fig:bmvs_compare}
\end{figure}

\subsection{Comparisons}

\noindent
\textbf{Quantitative comparisons.}
We perform quantitative comparisons with the SOTA methods on DTU dataset. We measure the Chamfer Distances of the predicted meshes with ground truth point clouds, and report the results in Table~\ref{tab: dtu_eval}. The results show that our method outperforms the SOTA methods by a large margin in both generic setting and per-scene optimization setting. 
Our results obtained by a per-scene fine-tuning with 10k iterations (20 mins) shows remarkable improvements than those of per-scene optimization methods.
Note that IDR~\cite{yariv2020multiview} needs extra object masks for per-scene optimization while others do not need object masks, and we provide the results of IDR for reference. 

We further perform a fine-tuning with 10k iterations for IBRNet and MVSNerf with the three input images. With the fine-tuning, the results of IBRNet are improved compared with its generic setting but still worse than our fine-tuned results.
MVSNerf fails to perform a fine-tuning with the three input images, therefore, no meaningful geometries are extracted.
Furthermore, we observe that MVSNerf usually needs more than 10 images to perform a successful fine-tuning, and thus the failure might be caused by the radiance ambiguity problem.



\begin{table}[!tp]
\caption{Ablation studies on DTU dataset.}
\begin{subtable}{.49\linewidth}
\centering
\begin{tabular}{cc|c}
\hline
Scheme & Setting & Chamfer dist. \\
\hline
Single volume  & \multirow{2}{*}{Generic} &  1.80\\
Cas. volumes & & 1.56\\
\hline
Single volume & \multirow{2}{*}{Fine-tuning} & 1.32 \\
Cas. volumes &  & 1.21\\
\hline
\end{tabular}
\caption{The usefulness of Cascaded volumes \\
in both generic and fine-tuning settings.}
\end{subtable}%
\begin{subtable}{.49\linewidth}
\centering
\begin{tabular}{ccc|c}
\hline
Pixel & Patch & Consistency & Chamfer dist. \\
\hline
$\checkmark$ & $\times$ & $\times$ & 1.39 \\
$\checkmark$ & $\checkmark$ & $\times$ & 1.28\\
$\checkmark$ & $\checkmark$ & $\checkmark$ & 1.21\\
\hline
\end{tabular}
\caption{The usefulness of Pixel-based and Patch-based blending, and Consistency-aware scheme in per-scene fine-tuning.}
\end{subtable} 
\label{tab: ablation}
\end{table}
\begin{figure}[!tp]
    \centering
  \includegraphics[width=\textwidth]{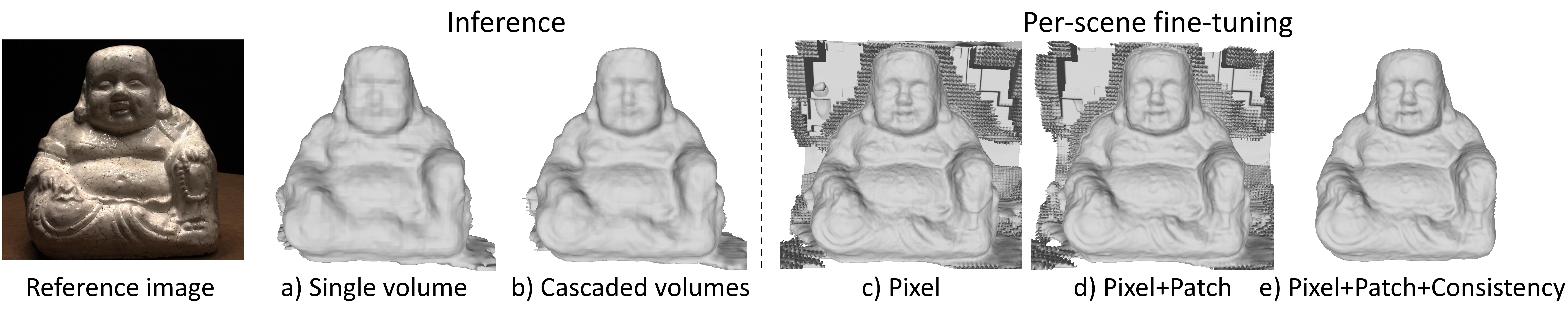}
    \caption{Qualitative ablation studies. The result obtained by cascaded volumes presents more fine-grained details than that of a single volume. The consistency-aware scheme can automatically detect the regions lacking radiance consistency and exclude them in the fine-tuning, thus yielding cleaner result (e) than the results without consistency-aware scheme (c,d). }
    \label{fig:ablation}
\end{figure}

\noindent
\textbf{Qualitative comparisons.}
We conduct qualitative comparisons with MVSNerf~\cite{chen2021mvsnerf}, COLMAP~\cite{schonberger2016structure} and NeuS~\cite{wang2021neus} on DTU~\cite{jensen2014large} and BlendedMVS~\cite{yao2020blendedmvs} datasets.
As shown in Figure~\ref{fig:dtu_compare}, our results obtained via network inference are much smoother and less noisy than those of MVSNerf.
The extracted meshes of MVSNerf are noisy since its representation of density implicit field does not have sufficient constraint on level sets of 3D geometry surfaces.

After a short-time per-scene fine-tuning, our results are largely improved with fine-grained details and become more accurate and cleaner.
Compared with the results of COLMAP, our results are more complete, especially for the objects with weak textures. With only three input images, NeuS suffers from the radiance ambiguity problem and its geometry surfaces are distorted and incomplete.

To validate the generalizability and robustness of our method, we further perform cross dataset evaluation on BlendedMVS dataset.
As shown in Figure~\ref{fig:bmvs_compare}, although our method is not trained on BlendedMVS, our generic model shows strong generalizability and produces cleaner and more complete results than those of MVSNerf. 
Take the fourth scene in Figure~\ref{fig:bmvs_compare} as an example, our method successfully recovers subtle details like the hose, while COLMAP misses the fine-grained geometry. For the scenes with weak textures, NeuS can only produces rough shapes and struggles to recover the details of geometry.

\subsection{Ablations and analysis}
\noindent
\textbf{Ablation studies.}
We conduct ablation studies (Table~\ref{tab: ablation} and Figure~\ref{fig:ablation}) to investigate the individual contribution of the important designs of our method.
The ablation studies are evaluated on one set of three images of the 15 testing scenes.
The first key module is a \textit{multi-level geometry reasoning scheme} for progressively constructing the surface from coarse to fine. Specially, a cascaded volume scheme is proposed, a coarse volume to generate coarse but high-level geometry, a fine volume guided by the coarse level to refine the geometry. 
As shown in (a) of Table~\ref{tab: ablation}, the cascaded volumes scheme considerably boosts the performance of our method than single volume scheme.
In Figure~\ref{fig:ablation}, we can see the geometry obtained by cascaded volumes contains more detailed geometry than that of a single volume.

The second important design is a multi-scale color blending strategy, which can enforce the local and contextual radiance consistency of rendered colors and ground truth colors.
As shown in (b) of Table~\ref{tab: ablation}, the combination of pixel-based and patch-based blending is better than solely using the pixel-based blending.
Another important strategy is a consistency-aware scheme that automatically detects the regions lacking photographic consistency and excludes these regions in fine-tuning.
As shown in (b) of Table~\ref{tab: ablation} and Figure~\ref{fig:ablation}, result using consistency-aware scheme is noticeably better than those that do not, which is cleaner and gets rid of distorted geometries.





\begin{wrapfigure}{r}{0.5\textwidth}
  \begin{center}
    \includegraphics[width=0.49\textwidth]{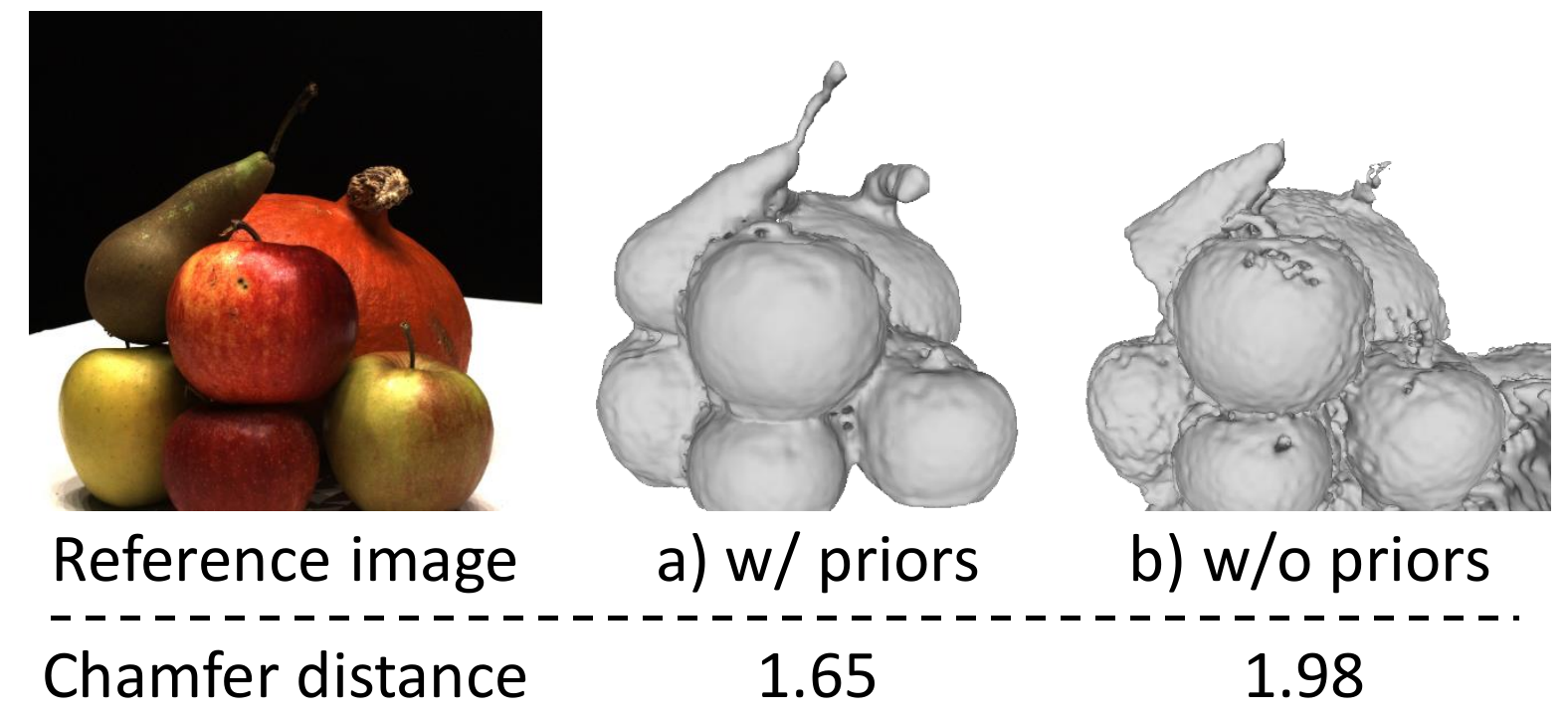}
  \end{center}
  \caption{Per-scene optimization with priors or without priors.}
    \label{fig:scratch}
\end{wrapfigure}

\noindent
\textbf{Per-scene optimization with or without priors.} 
Owing to the good initialization provided by the learned priors, the per-scene optimization of our method converges much faster and avoids sub-optimal caused by the radiance ambiguity problem.
To validate the effectiveness of the learned priors, we directly perform an optimization without using the learned priors.
As shown in Figure~\ref{fig:scratch}, the Chamfer Distance of the result with priors is 1.65 while that without prior-based initialization is 1.98. Obviously, the result with learned priors is more complete and smooth, which shows a stark contrast to the direct optimization.

\section{Conclusions}
We propose {\em SparseNeuS}, a novel neural rendering based surface reconstruction method to recover surfaces from multi-view images.
Our method generalizes to new scenes and produces high-quality reconstructions with sparse images, which prior works~\cite{wang2021neus,yariv2021volume,oechsle2021unisurf} struggle with.
To make our method generalize to new scenes, we therefore introduce \textit{geometry encoding} volumes to encode geometry information for generic geometry reasoning.
Moreover, a series of strategies are proposed to handle the difficult sparse views setting.
First, we propose a multi-level geometry reasoning framework to recover the surfaces in a coarse-to-fine manner. Second, we adopt a multi-scale color blending scheme, which jointly evaluates local and contextual radiance consistency for more reliable color prediction. Third, a consistency-aware fine-tuning scheme is used to control the inconsistent regions caused by occlusion and image noises, yielding accurate and clean reconstruction. 
Experiments show our method achieve better performance than the state-of-the-arts in both reconstruction quality and computational efficiency. 
Due to signed distance field adopted, our method can only produce closed-surfaces reconstructions.
Possible future directions include utilizing other representations like unsigned distance field to reconstruct open-surfaces objects.

\section*{Acknowlegements}
We thank the valuable feedbacks of reviewers. Xiaoxiao Long is supported by Hong Kong PhD Fellowship Scheme.

\clearpage
%
%
\bibliographystyle{splncs04}

\end{document}


\pagestyle{headings}
\mainmatter
\def\ECCVSubNumber{2516}  

\title{
Supplementary Materials for \\
SparseNeuS: Fast Generalizable Neural Surface Reconstruction from Sparse Views
} 

\titlerunning{Fast Generalizable Neural Surface Reconstruction from Sparse Views}
%
\author{Xiaoxiao Long\inst{1} \quad 
Cheng Lin\inst{2} \quad 
Peng Wang\inst{1} \\
Taku Komura\inst{1} \quad  Wenping Wang\inst{3}}
\authorrunning{X. Long et al.}
%
\institute{The University of Hong Kong \and Tencent Games \and Texas A\&M University
}

\newcommand{\Cheng}[1]{{\color{magenta} (Cheng: #1)}}
\newcommand{\wpeng}[1]{{\color{orange} (Peng: #1)}}

\maketitle


\definecolor{purple}{cmyk}{0.45,0.86,0,0}
\definecolor{bleudefrance}{rgb}{0.19, 0.55, 0.91}
\definecolor{darkorange}{rgb}{1, 0.55, 0}
\definecolor{limegreen}{rgb}{0.2, 0.8, 0.2}

\def\etal{et al.}			  
\def\eg{e.g.,~}               
\def\ie{i.e.,~}               
\def\etc{etc}                 
\def\cf{cf.~}                 
\def\viz{viz.~}               
\def\vs{vs.~}                 



\newlength\paramargin
\newlength\figmargin
\newlength\secmargin
\newlength\figcapmargin

\setlength{\secmargin}{0.0mm}
\setlength{\paramargin}{0.0mm}
\setlength{\figmargin}{0.0mm}
\setlength{\figcapmargin}{0.5mm}

\newcommand{\red}{\textcolor{red}}
\newcommand{\blue}{\textcolor{blue}}

\newcommand{\mpage}[2]
{
\begin{minipage}{#1\linewidth}\centering
#2
\end{minipage}
}

\newcommand{\mfigure}[2]
{
\begin{subfigure}[b]{#1\linewidth}\centering
\includegraphics[width=\linewidth]{#2}
\end{subfigure}
}

\newcommand{\Paragraph}[1]
{
\vspace{\paramargin}
\paragraph{#1}
}

\newcommand{\heading}[1]
{
\vspace{1mm}
\noindent \textbf{#1}
}   

\newcommand{\secref}[1]{Section~\ref{#1}}
\newcommand{\figref}[1]{Figure~\ref{#1}} 
\newcommand{\tblref}[1]{Table~\ref{#1}}
\newcommand{\eqnref}[1]{Equation~\ref{#1}}
\newcommand{\thmref}[1]{Theorem~\ref{#1}}
\newcommand{\prgref}[1]{Program~\ref{#1}}
\newcommand{\algref}[1]{Algorithm~\ref{#1}}
\newcommand{\clmref}[1]{Claim~\ref{#1}}
\newcommand{\lemref}[1]{Lemma~\ref{#1}}
\newcommand{\ptyref}[1]{Property~\ref{#1}}

\long\def\ignorethis#1{}
\newcommand {\todo}{{\textbf{\color{red}[TO-DO]\_}}}
\def\newtext#1{\textcolor{blue}{#1}}
\def\modtext#1{\textcolor{red}{#1}}

\newcommand{\LXX}[1]{{\color{cyan}          {[Xiao: #1]}}}
\newcommand{\ctc}[1]{{\color{darkorange}  {[CT: #1]}}}
\newcommand{\LC}[1]{{\color{blue}  {[LC: #1]}}}

\newcommand{\jbox}[2]{
  \fbox{%
  	\begin{minipage}{#1}%
  		\hfill\vspace{#2}%
  	\end{minipage}%
  }}

\newcommand{\jblock}[2]{%
	\begin{minipage}[t]{#1}\vspace{0cm}\centering%
	#2%
	\end{minipage}%
}

\section{Details of Patch-based Color Blending}
Besides the pixel-based color blending, we introduce patch-based color blending to jointly evaluate local and contextual radiance consistency, thus yielding more reliable color predictions.
To render the colors of a patch with size $k \times k$, we leverage local surface assumption and homography transformation for an efficient implementation.

The key idea is to estimate a local plane of a sampled point to efficiently derive the local patch. Given a sampled point $q$ in the query ray, we leverage the property of the SDF network $s(q)$ to estimate the normal direction $n_q$ by computing the spatial gradient, i.e., $n_q=\nabla s(q)$. Then, we select a set of points on the local plane $(q, n_q)$, project the selected points to each view, and obtain the colors by interpolation on each input image.
This projection operation is implemented by homography transformation.
Let $H$ be the homography between the view to be rendered $I_r$ and the $i_{th}$ input view $I_i$ induced by the local plane $(q, n_q)$:
\begin{equation}
H=K\left(R_{i}+\frac{{t}_{i} n_{q}^{T} R_{r}^{T}}{n_{q}^{T}\left(q+R_{r}^{T} t_{r}\right)}\right) K^{-1},
\end{equation}
where $K$ is the intrinsic matrix, $R_{i}$ is the $3 \times 3$ rotation matrix of $I_i$ relative to $I_r$,  $t_{i}$ is the 3D translation vector of $I_i$ relative to $I_r$, and $(R_{i}, t_{i})$ is the pose of the view $I_r$ in the world coordinate system.
Given the homography, we can obtain the projected pixel location $Hq$ on the view $I_i$, that is, the matrix product of $H$ and $q$, and then obtain $q$'s corresponding color by interpolation.

This homography is also applied to the set of points selected on the local plane $(q, n_q)$, so we can obtain their colors in the view $I_i$ by interpolation.
All the points on the local plane share the same blending weights with $q$, and thus only one query of the blending weights is needed.
By blending the patch colors interpolated from each view ${\{I_i\}}_{i=0}^{N-1}$ with the blending weights, we obtain the final patch colors of point $q$.
Same as pixel-based blending, we use SDF-based volume rendering~\cite{wang2021neus} to aggregate the interpolated patch colors of all the points sampled in the query ray $r$ to generate the final predicted patch colors of the ray.

Using local plane assumption, we consider the neighboring geometric information of a query 3D position, which encodes contextual information of local patches and enforces better geometric consistency. By adopting patch-based volume rendering, synthesized regions contain more global information than single pixels, thus producing more informative and consistent shape context, especially in the regions with weak texture and changing intensity.

\section{More Implementation Details}

\noindent
\textbf{Network details.}
Feature Pyramid Network~\cite{lin2017feature} is used as the image feature extraction network to extract multi-scale features from input images.
We implement the sparse 3D CNN networks using a U-Net like architecture, and use torchsparse~\cite{tang2020searching} as the implementation of 3D sparse convolution.
The signed distance function (SDF) $f_{\theta}$ is modeled by an MLP consisting of 4 hidden layers with a hidden size of 256. The blending network $f_{c}$ used in fine-tuning is modeled by an MLP consisting of 3 hidden layers with a hidden size of 256.
Positional encoding~\cite{mildenhall2020nerf} is applied to 3D locations with 6 frequencies and to view directions with 4 frequencies.
Same as NeuS~\cite{wang2021neus}, we adopt a hierarchical sampling strategy to sample points in the query ray for volume rendering, where the numbers of the coarse and fine sampling are both 64.

\noindent
\textbf{Training parameters.} The loss weights of total loss Eq.7 are set to $\alpha = 0.1, \beta=0.02$. The sdf scaling parameter $\tau$ of sparseness loss term Eq.10 is set to 100. For the consistency-aware color loss term Eq.6 used in fine-tuning, by default, $\lambda_0$ is set to 0.01 and $\lambda_{1}$ is set to 0.015. 
The ratio $\lambda_{0}/\lambda_{1}$ sometimes needs to be tuned for better reconstruction results for each scene: decreasing the ratio $\lambda_{0}/\lambda_{1}$ will lead to more regions being kept; otherwise, more regions are excluded and the surfaces are cleaner.

\noindent
\textbf{Data preparation.}
We observe that the images of the DTU dataset contain large black backgrounds and the regions have considerable image noises. Hence, we utilize a simple threshold-based denoising strategy to clean the images of training scenes.
We first detect the pixels where intensities are smaller than a threshold $\tau=10$ as the invalid black regions, and thus yielding a mask for each image. 
The mask is then processed by image dilation and erosion operations to reduce isolated outliers. 
Finally, we evaluate the areas of the connected components in the masks, and only keep the connected components whose areas are larger than $s$, where $s$ is set to the $10\%$ of the whole image.
Given the masks, the detected black invalid regions are set to 0.
By the simple denoising operation, the noises of the black background regions in the DTU training images are mostly removed.

\section{More experiments}
\begin{figure}
    \centering
  \includegraphics[width=\textwidth]{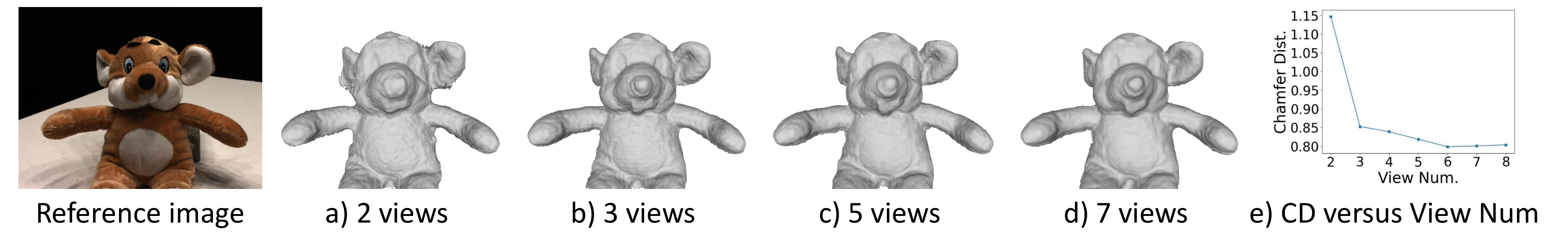}
    \caption{The results with different number of views as input.}
    \label{fig:views_num}
\end{figure}

\noindent
\textbf{Different number of views as input.}
Despite the good performance given the input of sparse images, our method can deal with an arbitrary number of input views. We investigate how the reconstruction quality is improved with more views as input. We conduct experiments on Scan105 of DTU dataset, with $2 \sim 8$ views as input, of which results are shown in Figure~\ref{fig:views_num}. Our method is still able to produce plausible geometries using only two views of an unseen object. With more views included, the reconstruction quality can also be progressively improved, and finally converges to a fairly low reconstruction error.

\begin{figure}[!htp]
    \centering
  \includegraphics[width=\textwidth]{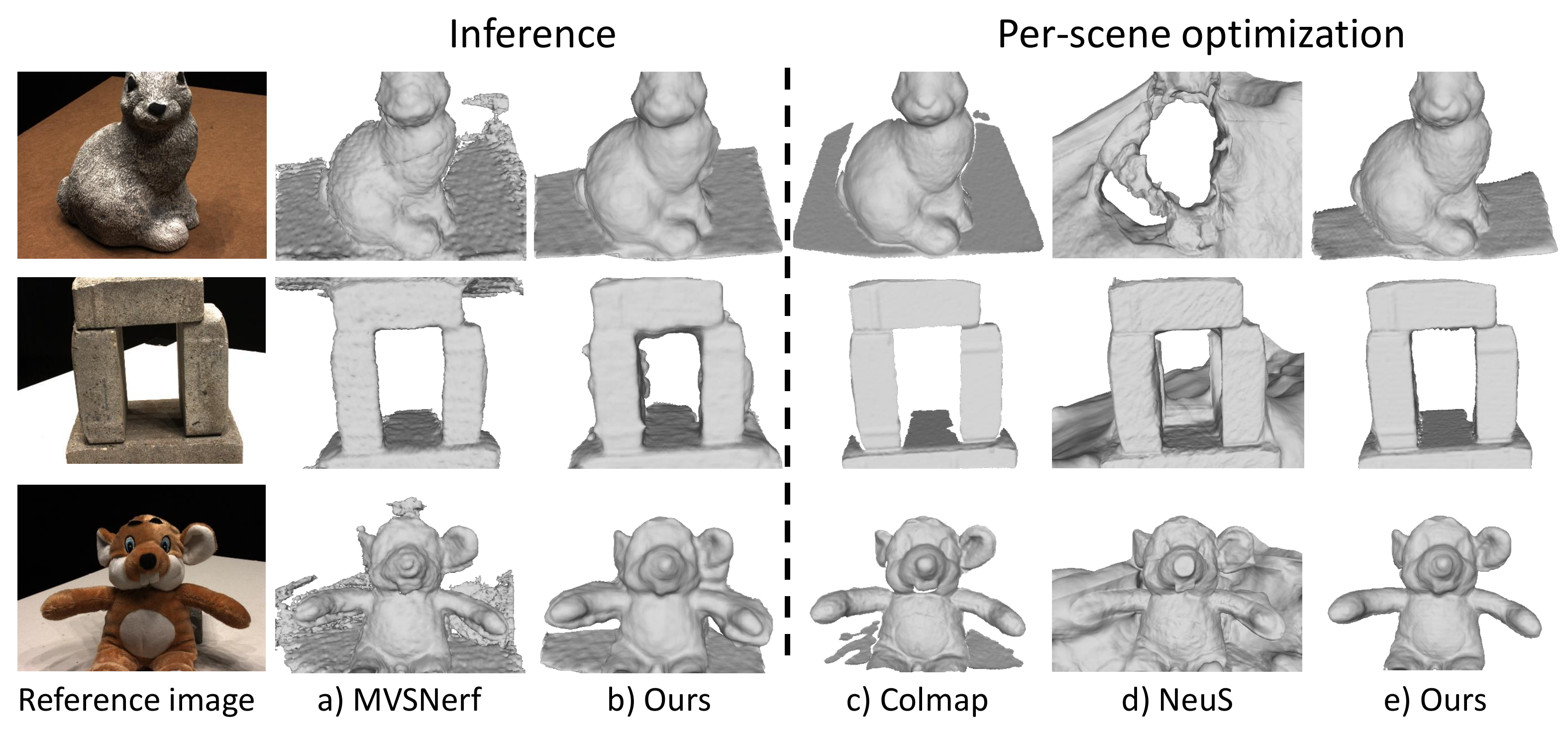}
    \caption{Visual comparisons on DTU~\cite{jensen2014large} dataset.}
    \label{fig:dtu_supp}
\end{figure}

\noindent
\textbf{More qualitative results.}
We present more qualitative comparisons with MVSNerf~\cite{chen2021mvsnerf}, Colmap~\cite{schonberger2016structure} and NeuS~\cite{wang2021neus} on DTU~\cite{jensen2014large} and BlendedMVS~\cite{yao2020blendedmvs} datasets.
As shown in Figure~\ref{fig:dtu_supp}, 
the extracted meshes of MVSNerf always suffer from noisy surfaces, while our results via fast network inference are much smoother and less noisy. This is because MVSNerf adopts density representation which lacks local surface constraint.

After a short-time per-scene fine-tuning, our results are noticeably improved with fine-grained details and become more accurate and cleaner.
Compared with the results of NeuS, our reconstructed surfaces are more complete and accurate. NeuS suffers from radiance ambiguity problem, and its geometries are incomplete and distorted.

\begin{figure}[!htp]
    \centering
  \includegraphics[width=\textwidth]{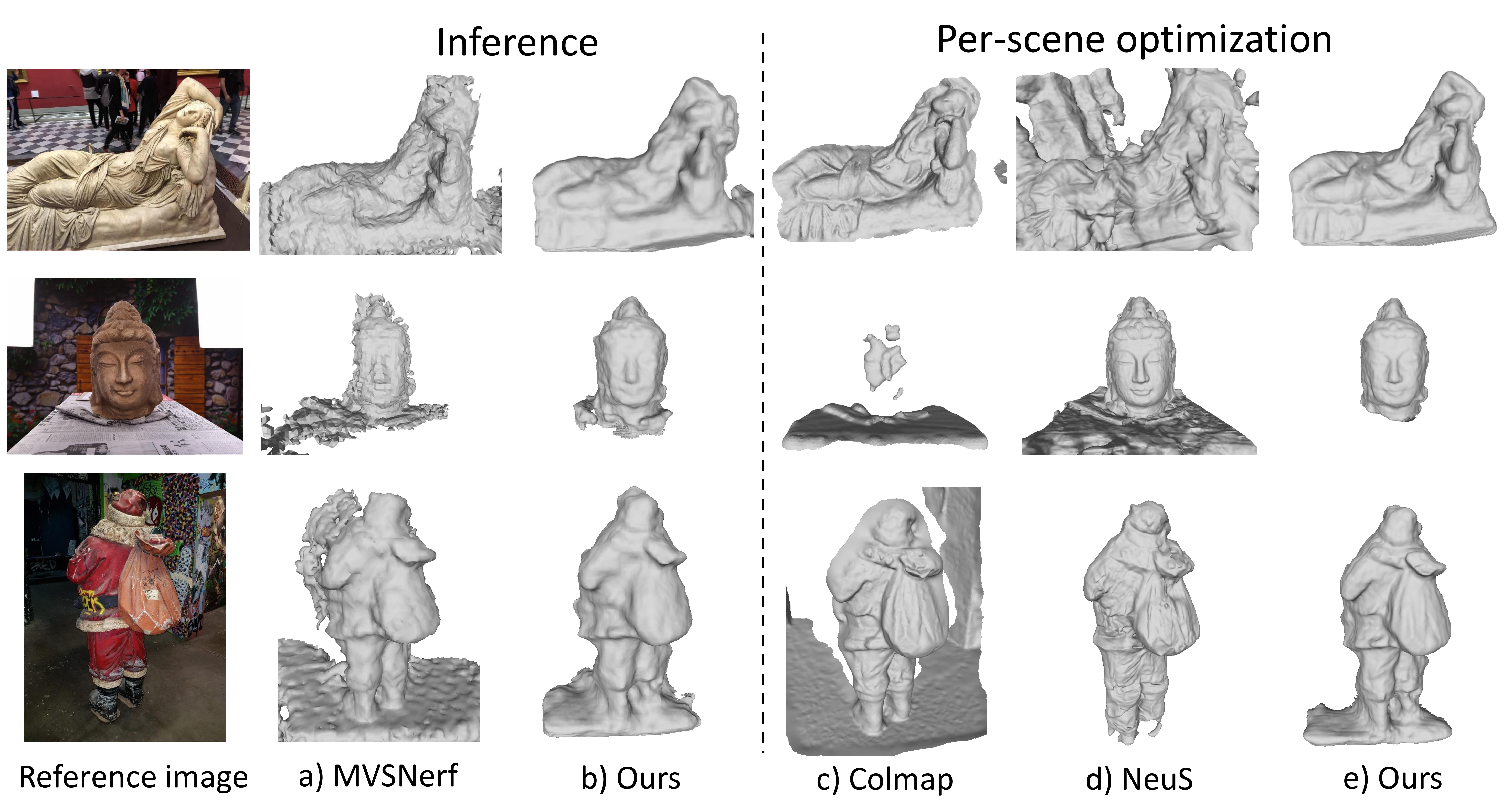}
    \caption{Visual comparisons on BlendedMVS~\cite{yao2020blendedmvs} dataset.}
    \label{fig:bmvs_supp}
\end{figure}

More comparisons on BlendedMVS dataset are presented in Figure~\ref{fig:bmvs_supp}.
Although our method is not trained on BlendedMVS, our generic model shows strong generalizability and produces cleaner and more complete results than those of MVSNerf. 
For example, for the Buddha head in Figure~\ref{fig:bmvs_supp}, Colmap fails to recover complete geometry and can only produce sparse points, while ours produces much more complete results.


\clearpage
%
%
\bibliographystyle{splncs04}
\bibliography{egbib}